# Stacked Bidirectional and Unidirectional LSTM Recurrent Neural Network for Network-wide Traffic Speed Prediction

Zhiyong Cui, Ruimin Ke, Ziyuan Pu, Yinhai Wang

*Abstract*— **Short-term traffic forecasting based on deep learning methods, especially long short-term memory (LSTM) neural networks, has received much attention in recent years. However, the potential of deep learning methods in traffic forecasting has not yet fully been exploited in terms of the depth of the model architecture, the spatial scale of the prediction area, and the predictive power of spatial-temporal data. In this paper, a deep stacked bidirectional and unidirectional LSTM (SBU-LSTM) neural network architecture is proposed, which considers both forward and backward dependencies in time series data, to predict network-wide traffic speed. A bidirectional LSTM (BDLSM) layer is exploited to capture spatial features and bidirectional temporal dependencies from historical data. To the best of our knowledge, this is the first time that BDLSTMs have been applied as building blocks for a deep architecture model to measure the backward dependency of traffic data for prediction. The proposed model can handle missing values in input data by using a masking mechanism. Further, this scalable model can predict traffic speed for both freeway and complex urban traffic networks. Comparisons with other classical and state-of-the-art models indicate that the proposed SBU-LSTM neural network achieves superior prediction performance for the whole traffic network in both accuracy and robustness.**

## I. INTRODUCTION

THE performances of intelligent transportation systems (ITS) applications largely rely on the quality of traffic information. Recently, with the significant increases in both the total traffic volume and the data they generate, opportunities and challenges exist in transportation management and research in terms of how to efficiently and accurately understand and exploit the essential information underneath these massive datasets. Short-term traffic forecasting based on data driven models for ITS applications has been one of the biggest developing research areas in utilizing massive traffic data, and has great influence on the overall performance of a variety of modern transportation systems [1].

In the last three decades, a large number of methods have been proposed for traffic forecasting in terms of predicting speed, volume, density and travel time. Studies in this area normally focus on the methodology components, aiming at developing different models to improve prediction accuracy, efficiency, or robustness. Previous literature indicates that the existing models can be roughly divided into two categories, i.e. classical statistical methods and computational intelligence (CI) approaches [2]. Most statistical methods for traffic forecasting were proposed at an earlier stage when traffic condition were less complex and transportation datasets were relatively small in size. Later on, with the rapid development in traffic sensing technologies and computational power, as well as traffic data volume, the majority of more recent work focuses on CI approaches for traffic forecasting.

With the ability to deal with high dimensional data and the capability of capturing non-linear relationship, CI approaches tend to outperform the statistical methods, such as auto-regressive integrated moving average (ARIMA) [36], with respect to handling complex traffic forecasting problems [38]. However, the full potential of artificial intelligence was not exploited until the rise of neural networks (NN) based methods. Ever since the precursory study of utilizing NN into the traffic prediction problem was proposed [39], many NN-based methods, like feed forward NN [41], fuzzy NN [40], recurrent NN (RNN) [42], and hybrid NN [25], are adopted for traffic forecasting problems. Recurrent Neural Networks (RNNs) model sequence data by maintaining a chain-like structure and internal memory with loops [4] and, due to the dynamic nature of transportation, are especially suitable to capture the temporal evolution of traffic status. However, the chain-like structure and the depth of the loops make RNNs difficult to train because of the vanishing or blowing up gradient problems during the back-propagating process. There have been a number of attempts to overcome the difficulty of training RNNs over the years. These difficulties were successfully addressed by the Long Short-Term Memory networks (LSTMs) [3], which is a type of RNN with gated structure to learn long-term dependencies of sequence-based tasks.

As a representative deep learning method handling sequence-

Zhiyong Cui, Ruimin Ke, Ziyuan Pu, and Yinhai Wang are with the Department of Civil and Environmental Engineering, University of Washington, Seattle, WA 98195 USA (e-mail: zhiyongc@uw.edu, ker27@uw.edu, ziyuanpu@uw.edu, yinhai@uw.edu).

data, LSTMs have been proved to be able to process sequence data [4] and applied in many real-world problems, like speech recognition [6], image captioning [7], music composition [8] and human trajectory prediction [9]. In recent years, LSTMs have been gaining popularity in traffic forecasting due to their ability to model long-term dependencies. Several studies [2, 22-25, 34, 43, 44, 45] have been done to examine the applicability of LSTMs in traffic forecasting, and the results demonstrate the advantages of LSTMs. However, the potential of LSTMs is far from being fully exploited in the domain of transportation. The three primary limitations in previous work on LSTMs in traffic forecasting can be summarized as follows: 1) traffic forecasting has generally focused on a small collection of network level. 2) Most of the structures of LSTM-based methods are shallow. 3) The long-term dependencies are normally learned from chronologically arranged input data considering only forward dependencies, while backward dependencies learned from reverse-chronological ordered data has never been explored.

From the perspective of the scale of prediction area, predicting large-scale transportation network traffic has become an important and challenging topic. Most existing studies utilize traffic data at a sensor location or along a corridor, and thus, network-wide prediction could not be achieved unless N models were trained for a traffic network with N nodes [22]. While, learning complex spatial-temporal features of a large-scale traffic network by only one model should be explored.

Regarding depth of the structure of LSTM-based models, the structure should have the ability to capture the dynamic nature of the traffic system. Most of the newly proposed LSTM-based prediction models have relatively shallow structures with only one hidden layer to deal with time series data [2, 22, 44]. Existing studies [20, 21] have shown that deep LSTM architectures with several hidden layers can build up progressively higher levels of representations of sequence data. Although some studies [23-25] utilized more than one hidden LSTM layer, the influences of the number of LSTM layers in different LSTM-based models need to be further compared and explained.

In terms of the dependency in prediction problems, all of the information contained in time series data should be fully utilized. Normally, the dataset fed to an LSTM model is chronologically arranged, with the result that the information in the LSTMs is passed in a positive direction from the time step $t-1$ to the time step $t$ along the chain-like structure. Thus, the LSTM structure only makes use of the forward dependencies [5]. But in this process, it is highly possible that useful information is filtered out or not efficiently passed through the chain-like gated structure. Therefore, it may be informative to consider backward dependencies, which pass information in a negative direction, into consideration. Another reason for including backward dependency into our study is the periodicity of traffic. Unlike wind speed forecasting [15], traffic incident forecasting [16], or many other time series forecasting problems with strong randomness, traffic conditions have strong periodicity and regularity, and even short-term periodicity can be observed [17]. Analysing the periodicity of time series data, especially for recurring traffic patterns, from both forward and backward temporal perspectives will enhance the predictive performance [28]. However, based on our review of the literature, few studies on traffic analysis utilized the backward dependency. To fill this gap, a bidirectional LSTM (BDLSTM) with the ability to deal with both forward and backward dependencies is adopted as a component of the network structure in this study.

In addition, when predicting the network-wide traffic speed, rather than the speed at a single location, the impact of upstream and downstream speeds on each location in the traffic network should not be neglected. Previous studies [26, 27] which only making use of the forward dependencies of time series data have found that the past speed values of upstream as well as downstream locations influence the future speed values of a location along a corridor. However, for complicated traffic networks with intersections and loops, upstream and downstream both refer to relative positions, and two arbitrary locations can be upstream and downstream of each other. Upstream and downstream are defined with respect to space, while forward and backward dependencies are defined with respect to time. With the help of forward and backward dependencies of spatial-temporal data, the learned feature will be more comprehensive.

In this paper, we propose a stacked bidirectional and unidirectional LSTM (SBU-LSTM) neural network, combining LSTM and BDLSTM, for network-wide traffic speed prediction. The proposed model is capable of handling input data with missing values and is tested on both large-scale freeway and urban traffic networks in the Seattle area. Experimental results show that our model achieves network-wide traffic speed prediction with a high prediction accuracy. The influence of the number of layers, the number of time lags (the length of time series input), the dimension of weight matrices in LSTM/BDLSTM layers, and the impact of additional volume and occupancy data are further analysed. The model's scalability and its potential applications are also discussed. In summary, our contributions can be stated as follows: 1) we expand the traffic forecasting area from a specific location or several adjacent locations along a corridor to large-scale traffic networks, varying from freeway traffic network to complex urban traffic network; 2) we propose a deep architecture considering backward dependencies by combining LSTM and BDLSTM to enhance the feature learning from the large-scale spatial time series data; 3) a masking mechanism is adopted to handle missing values; and 4) we evaluate many of the model's internal and external influential factors.

## II. METHODOLOGY

In this section, the components and the architecture of the proposed SBU-LSTM is detailly introduced in this section. Here, speed prediction is defined as predicting future speed based on historical speed information. The illustrations of the models in following sub-sections all take the traffic speed prediction as examples.

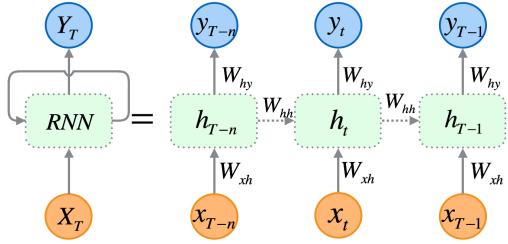

Fig. 1 Standard RNN architecture and an unfolded structure with T time steps

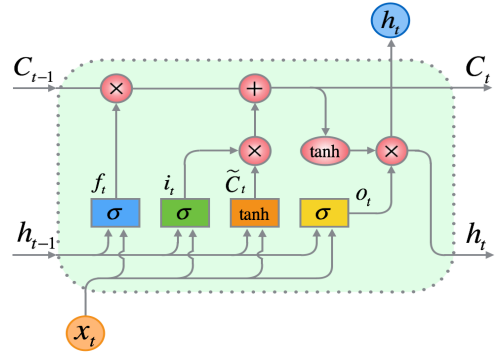

Fig. 2 LSTM architecture. The pink circles are arithmetic operators and the colored rectangles are the gates in LSTM.

### A. Network-wide Traffic Speed Data

Traffic speed prediction at one location normally uses a sequence of speed values with $n$ historical time steps as the input data [2, 22, 23], which can be represented by a vector,

$$X_T = [x_{T-n}, x_{T-(n-1)}, ..., x_{T-2}, x_{T-1}] \quad (1)$$

But the traffic speed at one location may be influenced by the speeds of nearby locations or even locations faraway, especially when traffic jam propagates through the traffic network. To take these network-wide influences into account, the proposed and compared models in this study take the network-wide traffic speed data as the input. Suppose the traffic network consists of P locations and we need to predict the traffic speeds at time T using n historical time frames (steps), the input can be characterized as a speed data matrix,

$$X_T^P = \begin{bmatrix} x^1 \\ x^2 \\ \vdots \\ x^P \end{bmatrix} = \begin{bmatrix} x_{T-n}^1 & x_{T-n+1}^1 & \cdots & x_{T-2}^1 & x_{T-1}^1 \\ x_{T-n}^2 & x_{T-n+1}^2 & \ddots & x_{T-2}^2 & x_{T-1}^2 \\ \vdots & \vdots & & \vdots & \vdots \\ x_{T-n}^P & x_{T-n+1}^P & \cdots & x_{T-2}^P & x_{T-1}^P \end{bmatrix} \quad (2)$$

where each element $x_t^p$ represents the speed of the $t$-th time frame at the $p$-th location. To reflect the temporal attributes of the speed data and simplify the expressions of the equations in the following subsections, the speed matrix is represented by a vector, $X_T^P = [x_{T-n}, x_{T-(n-1)}, ..., x_{T-2}, x_{T-1}]$, in which each element is a vector of the $P$ locations' speed values.

### B. RNNs

RNN is a class of powerful deep neural network using its internal memory with loops to deal with sequence data. The architecture of RNNs, which also is the basic structure of LSTMs, is illustrated in Fig. 1. For a hidden layer in RNN, it receives the input vector, $X_T^P$, and generates the output vector, $Y_T$. The unfolded structure of RNNs, shown in the right part of Fig. 1, presents the calculation process that, at each time iteration, $t$, the hidden layer maintains a hidden state, $h_t$, and updates it based on the layer input, $x_t$, and previous hidden state, $h_{t-1}$, using the following equation:

$$h_t = \sigma_h(W_{xh}x_t + W_{hh}h_{t-1} + b_h) \quad (3)$$

where $W_{xh}$ is the weight matrix from the input layer to the hidden layer, $W_{hh}$ is the weight matrix between two consecutive hidden states ($h_{t-1}$ and $h_t$), $b_h$ is the bias vector of the hidden layer and $\sigma_h$ is the activation function to generate the hidden state. The network output can be characterized as:

$$y_t = \sigma_y(W_{hy}h_t + b_y) \quad (4)$$

where $W_{hy}$ is the weight matrix from the hidden layer to the output layer, $b_y$ is the bias vector of the output layer and $\sigma_y$ is the activation function of the output layer. By applying the

Equation (1) and Equation (2), the parameters of the RNN is trained and updated iteratively via the back-propagation (BP) method. In each time step $t$, the hidden layer will generate a value, $y_t$, and the last output, $y_T$, is the desired predicted speed in the next time step, namely $\hat{x}_{T+1} = y_T$.

Although RNNs exhibit the superior capability of modeling nonlinear time series problems [2], regular RNNs suffering from the vanishing or blowing up gradient during the BP process, and thus, being incapable of learning from long time lags [10], or saying long-term dependencies [11].

### C. LSTMs

To handle the aforementioned problems of RNNs, several sophisticated recurrent architectures, like LSTM architecture [3] and Gated Recurrent Unit (GRU) architecture [12] are proposed. It has been showed that the LSTMs work well on sequence-based tasks with long-term dependencies, but GRU, a simplified LSTM architecture, is only recently introduced and used in the context of machine translation [13]. Although there are a variety of typical LSTM variants proposed in recent year, a large-scale analysis of LSTM variant shows that none of the variants can improve upon the standard LSTM architecture significantly [14]. Thus, the standard LSTM architecture is adopted in this study as a part of the proposed network structure and introduced in this section.

The only different component between standard LSTM architecture and RNN architecture is the hidden layer [10]. The hidden layer of LSTM is also named as LSTM cell, which is shown in Fig. 2. Like RNNs, at each time iteration, $t$, the LSTM cell has the layer input, $x_t$, and the layer output, $h_t$. The complicated cell also takes the cell input state, $\widetilde{C_t}$, the cell output state, $C_t$, and the previous cell output state, $C_{t-1}$, into account while training and updating parameters. Due to the gated structure, LSTM can deal with long-term dependencies to allow useful information pass along the LSTM network. There are three gates in a LSTM cell, including an input gate, a forget gate, and an output gate. The gated structure, especially the forget gate, helps LSTM to be an effective and scalable model for several learning problems related to sequential data [14]. At time $t$, the input gate, the forget gate and the output gate, denoted as $i_t$, $f_t$, and $o_t$ respectively. The input gate, the forget gate, the output gate and the input cell state, which are represented by colorful boxes in the LSTM cell in Fig. 2, can be calculated using the following equations:

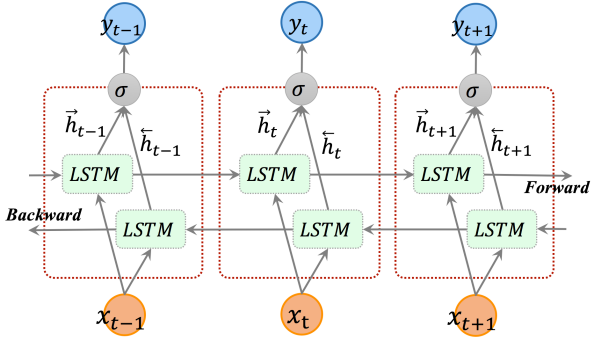

Fig. 3 Unfolded architecture of bidirectional LSTM with three consecutive steps

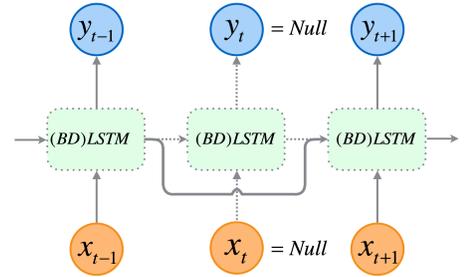

Fig. 4 Masking layer for time series data with missing values

$$f_t = \sigma_g(W_f x_t + U_f h_{t-1} + b_f) \tag{5}$$

$$i_t = \sigma_g(W_i x_t + U_i h_{t-1} + b_i) \tag{6}$$

$$o_t = \sigma_g(W_o x_t + U_o h_{t-1} + b_o) \tag{7}$$

$$\tilde{C}_t = \tanh(W_C x_t + U_C h_{t-1} + b_C) \tag{8}$$

where $W_f$, $W_i$, $W_o$, and $W_C$ are the weight matrices mapping the hidden layer input to the three gates and the input cell state, while the $U_f$, $U_i$, $U_o$, and $U_C$ are the weight matrices connecting the previous cell output state to the three gates and the input cell state. The $b_f$, $b_i$, $b_o$, and $b_C$ are four bias vectors. The $\sigma_g$ is the gate activation function, which normally is the sigmoid function, and the tanh is the hyperbolic tangent function. Based on the results of four above equations, at each time iteration $t$, the cell output state, $C_t$, and the layer output, $h_t$, can be calculated as follows:

$$C_t = f_t * C_{t-1} + i_t * \tilde{C}_t \tag{9}$$

$$h_t = o_t * \tanh(C_t) \tag{10}$$

The final output of a LSTM layer should be a vector of all the outputs, represented by $Y_T = [h_{T-n}, ..., h_{T-1}]$. Here, when taking the speed prediction problem as an example, only the last element of the output vector, $h_{T-1}$, is what we want to predict. Thus, the predicted speed value ($\hat{x}$) for the next time iteration, $T$, is $h_{T-1}$, namely $\hat{x}_T = h_{T-1}$.

### D. BDLSTMs

The idea of BDLSTMs comes from bidirectional RNN [18], which processes sequence data in both forward and backward directions with two separate hidden layers. BDLSTMs connect the two hidden layers to the same output layer. It has been proved that the bidirectional networks are substantially better than unidirectional ones in many fields, like phoneme classification [19] and speech recognition [20]. But bidirectional LSTMs have not been used in traffic prediction problem, based on our review of the literature [2,22,23,24,25].

In this section, the structure of an unfolded BDLSTM layer, containing a forward LSTM layer and a backward LSTM layer, is introduced and illustrated in Fig. 3. The forward layer output sequence, $\vec{h}$, is iteratively calculated using inputs in a positive sequence from time $T - n$ to time $T - 1$, while the backward layer output sequence, $\overleftarrow{h}$, is calculated using the reversed inputs from time $T - n$ to $T - 1$. Both the forward and backward layer outputs are calculated by using the standard LSTM updating equations, Equations (3) - (8). The BDLSTM layer generates an output vector, $Y_T$, in which each element is calculated by using the following equation:

$$y_t = \sigma(\vec{h}_t, \overleftarrow{h}_t) \tag{11}$$

where $\sigma$ function is used to combine the two output sequences. It can be a concatenating function, a summation function, an average function or a multiplication function. Similar to the LSTM layer, the final output of a BDLSTM layer can be represented by a vector, $Y_T = [y_{T-n}, ..., y_{T-1}]$, in which the last element, $y_{T-1}$, is the predicted speed for the next time iteration when taking speed prediction as an example.

### E. Masking Layer for Time Series Data with Missing Values

In reality, traffic sensors, like inductive-loop detectors, may fail due to breakdown of wire insulation, poor sealants, damage caused by construction activities, or electronics unit failure. The sensor failure further causes missing values in collected time series data. For the LSTM-based prediction problem, if the input time series data contains missing/null values, the LSTM-based model will fail due to null values cannot be computed during the training process. If the missing values are set as zero, or some other pre-defined values, the training and testing results will be highly biased. Thus, we adopt a masking mechanism to overcome the potential missing values problem.

Fig. 4 demonstrates the details of the masking mechanism. The (BD)LSTM cell denotes a LSTM-based layer, like a LSTM layer or a BDLSTM layer. A mask value, $\emptyset$, is pre-defined, which normally is 0 or Null, and all missing values in the time series data are set as $\emptyset$. For an input time series data $X_T$, if $x_t$ is the missed element, which equals to $\emptyset$, the training process at the $t$-th step will be skipped, and thus, the calculated cell state of the $(t - 1)$-th step will be directly input into the $(t + 1)$-th step. In this case, the output of $t$-th step also equals to $\emptyset$, which will be considered as a missing value and, if needed, input to the subsequent layer. Similarly, we can deal with input data with consecutive missing values using the masking mechanism.

### F. Stacked Bidirectional and Unidirectional LSTM Networks

Existing studies [20, 21] have shown that deep LSTM architectures with several hidden layers can build up progressively higher level of representations of sequence data, and thus, work more effective. The deep LSTM architectures are networks with several stacked LSTM hidden layers, in which the output of a LSTM hidden layer will be fed as the input into the subsequent LSTM hidden layer. This stacked-layers mechanism, which can enhance the power of neural networks, is adopted in this study. As mentioned in previous sections, BDLSTMs can make use of both forward and backward dependencies. When feeding the spatial-temporal information of the traffic network to the BDLSTMs, both the spatial correlation of the speeds in different locations of the traffic network and the temporal dependencies of the speed values can be captured during the feature learning process. In this regard, the BDLSTMs are very suitable for being the first layer of a model to learn more useful information from spatial time series data. When predicting future speed values, the top layer of the architecture only needs to utilize learned features, namely the outputs from lower layers, to calculate iteratively along the forward direction and generate the predicted values. Thus, an LSTM layer, which is fit for capturing forward dependency, is a better choice to be the last (top) layer of the model.

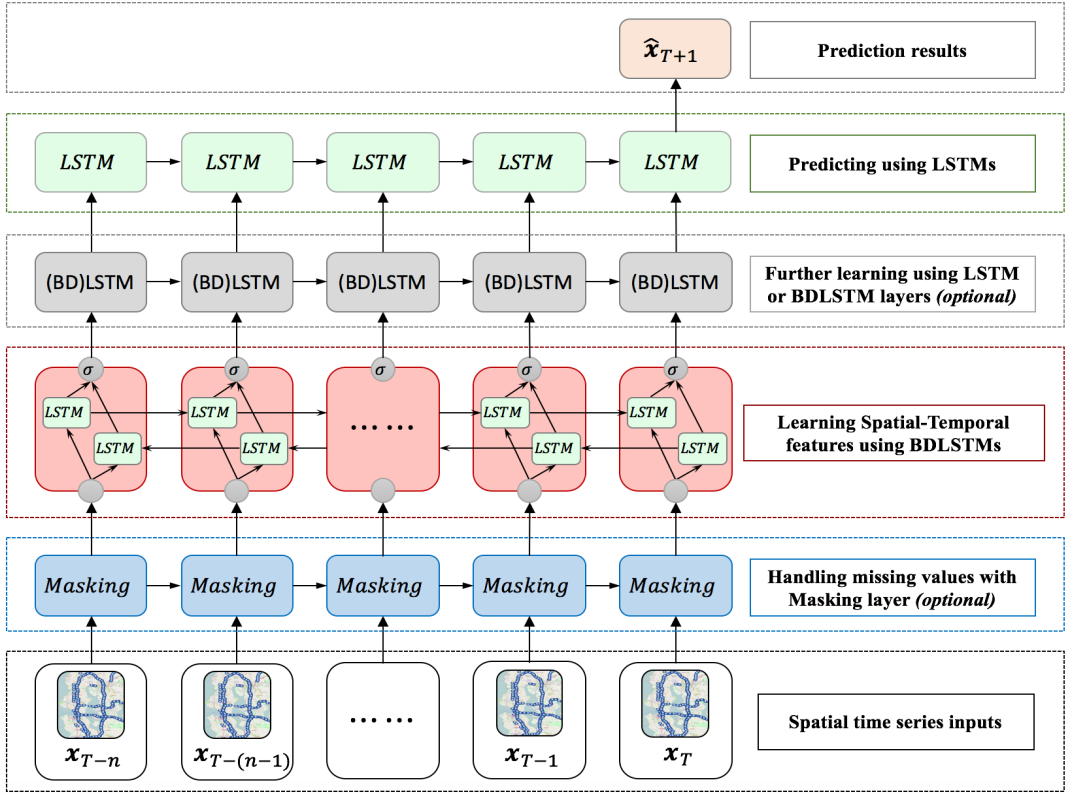

Fig. 5 SBU-LSTMs architecture necessarily consists of a BDLSTM layer and a LSTM layer. Masking layer for handling missing values and multiple LSTM or BDLSTM layers as middle layers are optional.

In this study, we propose a novel deep architecture named stacked bidirectional and unidirectional LSTM network (SBU-LSTM) to predict the network-wide traffic speed values. Fig. 5 illustrates the graphical architecture of the proposed model. If the input contains missing values, a masking layer should be adopted by the SBU-LSTM. Each SBU-LSTM contains a BDLSTM layer as the first feature-learning layer and a LSTM layer as the last layer. For sake of making full use of the input data and learning complex and comprehensive features, the SBU-LSTM can be optionally filled with one or more LSTM/BDLSTM layers in the middle. Fig. 5 shows that the SBU-LSTM takes the spatial time series data as the input and predict future speed values for one time-step. The SBU-LSTM is also capable of predicting values for multiple future time steps based on historical data. But this property is not shown in

Fig. 5, since the target of this study is to predict network-wide traffic speed for one future time step. The detailed spatial structure of input data is described in the experiment section.

## III. EXPERIMENTS

### A. Dataset Description

In this study, two types of traffic state datasets are utilized to carry out experiments to test the proposed model. One is a station-/point-based dataset, called loop detector data [46], collected by inductive loop detectors deployed on roadway surface. Multiple loop detectors are connected to a detector station deployed around every half a mile. The collected data from each station are grouped and aggregated as station-based traffic state data according to directions. This aggregated and quality controlled dataset contains traffic speed, volume, and occupancy information. In the experiments, the loop detector data cover four connected freeways, which are I-5, I-405, I-90 and SR-520 in the Seattle area, and are extracted from the Digital Roadway Interactive Visualization and Evaluation Network (DRIVE Net) system [29, 30]. The traffic sensor stations are shown in Fig. 6 (a), which is represented by small blue icons. This dataset contains traffic state data of 323 sensor stations in 2015 and the time step interval of this dataset is 5 minutes.

The other dataset used in this study is a segment-based dataset, called INRIX data [46], which measures traffic speeds of both freeway and urban roadway segments. INRIX data is selected by the U.S. Federal Highway Administration as the National Performance Management Research Data Set. INRIX data provide wide coverage and accurate traffic information by aggregating GPS probe data from a wide array of commercial vehicle fleets, connected cars and mobile apps. An entire traffic network in the Seattle downtown area, which contains more than 1000 roadway segments, shown in Fig. 6 (b), is selected as the experimental dataset. The dataset covers the whole year of 2012 and its time step interval is also 5 minutes.

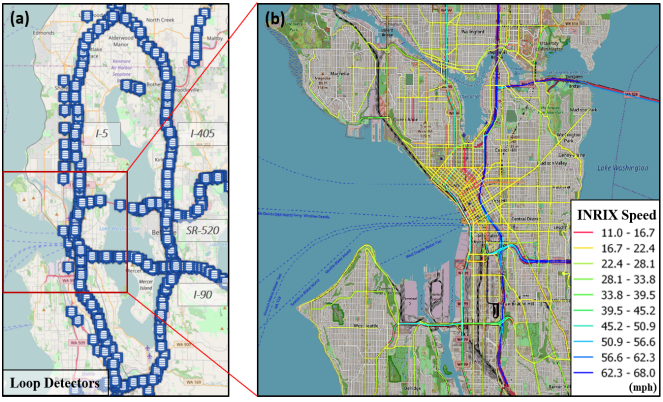

Fig. 6 Loop detector stations on the freeway network in Seattle area

| Models | MAE (mph) | MAPE (%) |
|---|---|---|
| SVM | 9.23 | 20.39 |
| Random Forest | 2.64 | 6.30 |
| Feed-forward NN (2-hidden layers) | 2.63 | 6.41 |
| GRU NN | 3.43 | 8.02 |
| SBU-LSTMs | 2.42 | 5.67 |

### B. Experiment Results Analysis and Comparison

In this sub-section, only the loop detector data, due to its high data quality [46], are used to measure the performance of the proposed approach and compare with other models. Hence, the network-wide traffic is characterized by the 323 station speed values and the spatial dimension of the input data is set as, $P = 323$. Since, the unit of a time step in loop detector data is 5 minutes, the dataset has $\frac{60(min)}{5} \times 24(hour) \times 365(year) = 105120$ time steps in total. Suppose the number of the time lags is set as $n = 10$, which means the model uses a set of data with 10 consecutive time steps (covering 50 minutes) to predict the following 5-minute speed value, the dataset is separated into samples with 10 time lags and the sample size is $N = 105110$ ($105120 - 10$).

Based on the descriptions of the model, each sample of the input data, $X_T^P$, is a 2-D vector with the dimension of $[n, P] = [10, 323]$, and each sample of the output data is a 1-dimension vector with 323 components. The input of the model is a 3-D vector, whose dimension is $[N, n, P]$. Before fed into the model, all the samples are randomized and divided into training set, validation set, and test set with the ratio 7:2:1.

In the training process, mini-batch gradient descent method is used when the model optimizes the mean squared error (MSE) loss using RMSProp optimizer and early stopping mechanism is used to avoid over-fitting. To measure the effectiveness of different traffic speed prediction algorithms, the Mean Absolute Errors (MAE) and Mean Absolute Percentage Errors (MAPE) are computed using the following equations:

$$MAE = \frac{1}{n} \sum_{i=1}^{n} |x_i - \hat{x}_i| \qquad (12)$$

$$MAPE = \frac{100}{n} \sum_{i=1}^{n} \left| \frac{x_i - \hat{x}_i}{x_i} \right| \qquad (13)$$

where $x_i$ is the observed traffic speed, and $\hat{x}_i$ is the predicted speed. All the compared models in this section are trained and tested multiple times to eliminate outliers, and the results of them presented are averaged to reduce random errors.

In this section, the results of the proposed SBU-LSTMs are analyzed and compared with classical methods and other RNN-based models. Further analysis about the influence of the number of time lags, the dimension of weight matrices, the number of layers, the impact of volume and occupancy information, spatial feature learning, and model robustness are carried out to shed more light on the characteristics of proposed model.

#### 1) Comparison with Classical Models for Single Location Traffic Speed Prediction

Many classical baseline models used in traffic forecasting problems, like ARIMA [2, 23] Support Vector Regression (SVR) [37], Kalman filter [35]. Based on our literature review [2], the performances of ARIMA and Kalman filter method are far behind the others, and thus, these two methods are not compared in this study. Most of mentioned classical models are not suitable for predicting network-wide traffic speed via a single model, since they normally cannot process 3-D spatial temporal vectors. To compare our proposed model with these baseline models, experiments are carried out for single loop detector stations, whose input data is a 2-D vector without spatial dimension. The results are averaged to measure the overall performance of these models.

We compared the performance of the SBU-LSTMs with SVR, random forest, feed-forward NN, GRU NN. In this comparison, the proposed model does not use the masking layer and optional middle layers. Among these baseline models, the feed-forward NN model, also called Multilayer Perceptron (MLP), has superior performance for the traffic flow prediction [32], and decision tree and SVR are very efficient models for prediction [23, 37]. For the SVR method, the Radial Basis Function (RBF) kernel is utilized, and for the Random Forest method, 10 trees are built, and no maximum depth of the trees is limited. In this experiment, the feed-forward NN model consists of two hidden layers with 323 nodes in each layer.

Table I demonstrates the prediction performance of different algorithms for the single detector stations. The number of input time lags in this experiment is set as 10. Among the non-neural network algorithms, random forest performs much better, with the MAE of 2.64, than the SVM method, which makes sense due to the majority votes mechanism of random forest. The feed-forward NN whose MAE is 2.63 performs very close to the random forest method. Although GRU NN is a kind of recurrent NN, its performance obviously cannot outperform those of feed-forward NN and random forest. The single layer structure and the simplified gates in GRU NN may be the reasons. To sum up, the proposed SBU-LSTM model is clearly superior to the other four methods in this single detector station based experiment.

TABLE II
PERFORMANCE COMPARISON OF THE PROPOSED MODEL WITH OTHER LSTM-BASED MODELS FOR NETWORK-WIDE TRAFFIC SPEED PREDICTION

| Model | Number of LSTM / BDLSTM layers | | | | | | | | | |
|---|---|---|---|---|---|---|---|---|---|---|
| | N = 0 | | N = 1 | | N = 2 | | N = 3 | | N = 4 | |
| | MAE | MAPE | MAE | MAPE | MAE | MAPE | MAE | MAPE | MAE | MAPE |
| **N**-layers LSTM | | | 2.886 | 6.585 | 2.502 | 5.929 | 2.483 | 5.950 | 2.529 | 6.114 |
| **N**-layers LSTM + 1-layer DNN | | | 2.652 | 6.489 | 2.581 | 6.332 | 2.630 | 6.438 | 2.646 | 6.586 |
| **N**-layers LSTM + Hour of Day + Day of Week | | | 2.668 | 6.506 | 2.557 | 6.274 | 2.595 | 6.447 | 2.647 | 6.602 |
| **N**-layers BDLSTM | | | 3.021 | 6.758 | 2.472 | 5.819 | 2.476 | 5.846 | 2.526 | 5.988 |
| SBU-LSTMs: 1-layer BDLSTM + N middle BDLSTM layers + 1-layer LSTM | **2.426** | **5.674** | 2.465 | 5.787 | 2.502 | 5.950 | 2.549 | 6.191 | 2.576 | 6.227 |

### 2) Comparison with LSTM-based models for Network-wide Traffic Speed Prediction

The SBU-LSTMs is proposed aiming at predicting the network-wide traffic speed, and thus, other methods with the ability of predicting multi-dimensional time series data are compared in this section. Since the proposed model combines BDLSTMs and LSTMs, the pure deep (N-layers) BDLSTMs and LSTMs are compared. A deep LSTM NN adding a fully connected deep neural network (DNN) layer, which is proven to be able to boost the LSTM NN [33], is also compared. To measure the influence of temporal information to the network-wide traffic speed, a multilayer LSTM model combining day of week and hour of day is also tested in this experiment.

Meanwhile, the influence of depth of the neural networks, namely the number of layers of the models, is tested in this section. All the experiments undertook in this section used the dataset covering the whole traffic network with 10-time lags. The number of time lags, 10, is set within a reasonable range for traffic forecasting based on literatures [25, 32] and our experiments. The spatial dimension of weight matrices in each LSTM or BDLSTM layer in this experiment is set as the number of loop detector stations, 323, to ensure the spatial feature can be fully captured. The comparison results are averaged from multiple tests to remove random errors.

Table II shows the comparison results, where the headers on horizontal axis show the amount of the LSTM or BDLSTM layers owned by the models. In terms of the influence of depth of the neural network, all the compared models achieve their best performance when they have two layers and their performances have the same trends that the values of MAE and MAPE increase as the number of layers increases from two to four. Table II contains a special "(N=0)" column, denoting no middle layer, to represent the basic structure of the SBU-LSTM. The performance of SBU-LSTM is in conformity with the trends of the compared models that the MAE and MAPE increase as the number of layers rises from zero to four.

The proposed SBU-LSTM outperforms the others for all the layer numbers. When the SBU-LSTM has no middle layer, it achieves the best MAE, 2.426 mph, and MAPE, 5.674%. The test errors of multilayer LSTM NN and BD LSTM NN turn out to be larger than that of the proposed model. They achieve their best MAEs of 2.502 and 2.472, respectively, when they both have two layers. It should be noted that, for the one-layer case, the BDLSTM NN model gets the worst performance in our experiments shown in the Table II. It indicates that one-layer BDLSTM may be good enough for capturing features, but it is not satisfactory to predict the results. Except for the one-layer case, the model combining deep LSTM and DNN are not comparable with others. This test results show that adding DNN layers to deep LSTM cannot make improvements for the network-wide traffic prediction problem is consistent with the finding in a previous study [33]. The performance of the temporal information added multilayer LSTM is very close to that of the LSTM combined with DNN. Thus, incorporating the day of week and time of day features cannot improve the performance for this study. This is in accordance with the results of previous works [23, 24].

### 3) Influence of number of time lags

The number of time lags, $n$, is the temporal dimension of the input data, which may influence the performance of the proposed model. Fig. 6 shows the boxplot of the MAE versus the number of time lags, in which the spatial dimensions of all weight matrices are all set as $P = 323$. When the number of time lags equals 8, 10, and 12, the MAEs are very close, around 2.4. The deviations of these MAEs are relatively small. When the number of time lags is set as 6, the MAE is much higher, and the deviation is much larger than other cases. That means, given the 5-minute time step interval and the studied traffic network, input data with 6 time steps are not enough for the model to accurately predict network-wide traffic speed. To sum up, the number of time lags tends to influence the predictive performance, especially when the number is relatively small.

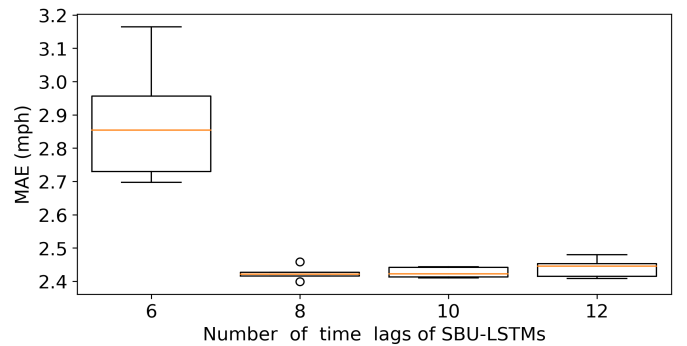

Fig. 6 Boxplot of MAE versus number of time lags in SBU-LSTMs. One unit of time lag is 5 minutes.

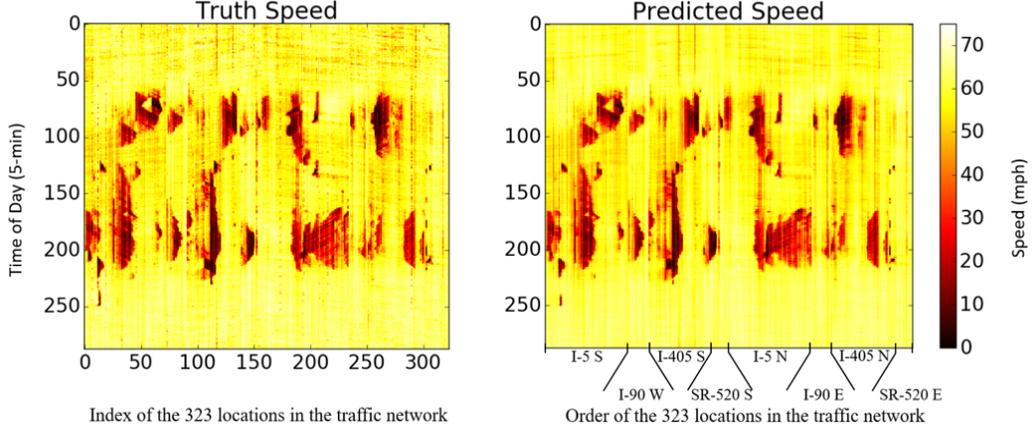

Fig. 7 Heatmaps of ground truth and predicted speed values for the freeway traffic network on 01/09/2015. The two plots share the same meanings of the two axes, where the two horizontal axes represent the index and the arrangement order of sensor stations based on the mileposts and directions of the four freeways, respectively.

### 4) Influence of dimension of weight matrices

In the experiment, the dimension of each data sample is $[n, P]$, where $P$ is the spatial dimension representing the number of loop detector stations. According to the matrix multiplication rule, the spatial dimension of the weight matrices in the first layer of the SBU-LSTM must be accordance with the value of $P$. But the spatial dimension of weight matrices in other layers can be customized. In this section, we measure the influence of the dimension of weight matrices in the basic SBU-LSTM.

When the model's last LSTM layer has different spatial dimensions, including $\left[\frac{1}{4}P\right]$, $\left[\frac{1}{2}P\right]$, $P$, $2P$ and $4P$, very close prediction results are observed. Here, $P$ equals 323 and $\lceil \cdot \rceil$ is the ceiling function. Table III shows the comparison results. The MAE, MAPE, and standard deviations are nearly the same. Hence, the variation of the dimension of the weight matrices in the LSTM layer almost has no influence on the predictive performance, if the dimension is set as a reasonable value close to the number of sensor locations.

### 5) Spatial features learning

Spatial features of a traffic network are critical for predicting network-wide traffic states. By carefully studying the LSTM methodology, we can find that the spatial features can be inherently learned by the weights in LSTM or BDLSTM layers at the training process. No matter what the network's spatial structure is, and no matter what the spatial order of the input data is, the traffic speed relationship between each pair of two locations in the traffic network can be captured by the LSTM weight matrices.

In this section, we measure the influence of spatial order of the input data on the spatial feature learning. Firstly, we order the spatial dimension of input data based on the milepost and direction of freeways. Fig. 7 displays the heatmap of true speed and predicted speed for the freeway network on a randomly selected day, taking 09/01/2015, a Friday, for an example. The extremely similarity between the shapes in the two heatmaps shows that the proposed model is capable of learning spatial features. Then, we randomly rearrange the spatial dimension of input data. By training and testing the model for multiple times, we find that the predictive performance nearly does not change, and the MAEs are all around 2.42 mph. To the best of our knowledge, at least two aspects of reasons lead the good performance. One is that the BDLSTM, measuring both forward and backward dependencies, helps learn better features. The other one is that the inherent spatial correlation between locations is learned and stored in the weight matrices during the training process. Hence, the order of spatial dimension of input data basically does not affect the model performance.

### 6) Influence of volume and occupancy

Speed, volume (flow), and occupancy are the three fundamental factors to analyze traffic flow. Considering the loop detector data contains speed, volume, and occupancy information, it is informative to investigate the influence of these factors on the proposed model's predictive performance. In previous experiments, each element of the model input, $x_t^p$, is the speed ($s$) at a specific location, $p$, at time $t$, where $x_t^p = s_t^p$. While, in this experiment, an element of the model input combine speed ($s$) with volume ($v$) and occupancy ($o$), where

TABLE III
PERFORMANCES COMPARISON OF SBU-LSTMs WITH DIFFERENT SPATIAL
DIMENSIONS OF WEIGHT MATRICES

| Spatial dimension of weight matrices in the last layer (LSTM layer) | MAE | MAPE | STD |
|---|---|---|---|
| $1/4\ P$ | 2.486 | 5.903 | 0.675 |
| $1/2\ P$ | 2.425 | 5.680 | 0.643 |
| $P = 323$ | 2.426 | 5.674 | 0.630 |
| $2\ P$ | 2.431 | 5.736 | 0.636 |
| $4\ P$ | 2.411 | 5.696 | 0.636 |

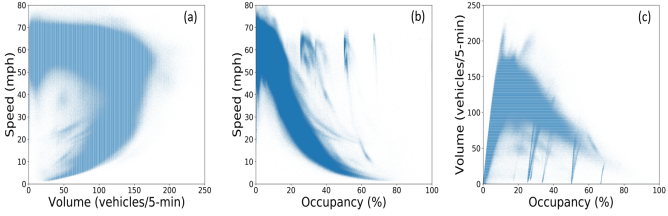

Fig. 8 Fundamental scatter diagrams of traffic flow: (a) speed-volume diagram, (b) speed-occupancy diagram, and (c) volume-occupancy diagram.

$x_t^p$ can be $(s, o)_t^p$, $(s, v)_t^p$, or $(s, v, o)_t^p$.

Before investigating the influence of volume and occupancy, the relationship between these factors are directly evaluated by three scatter diagrams, plotted based on the loop detector dataset and shown in Fig. 8. Although obvious noise points can be seen from Fig. 8 (b) and Fig. 8 (c), the main distributions in the three diagrams follows the traffic flow theory [47]. Our experiments show that when solely combining volume data, there is nearly no improvement over the prediction accuracy shown in the Section *2)*. But when inputting speed and occupancy, or all the three factors, the model performs slightly better, which has less than 5% increase in the prediction accuracy. Therefore, the volume and occupancy has slightly influence on the traffic speed prediction based on our experiment results.

### 7) Model robustness

The optional masking layer makes the SBU-LSTM more robust that the model can handle input data with missing values. In this section, we test the model's robustness by randomly selecting a specific proportion of elements from the spatial-temporal matrix in each input sample and set them as *Null*. The model's prediction accuracy varies as we change the proportion of *Null* values in the input samples.

The performance of the model is presented in Table IV. It is obvious that the prediction accuracy decreases as the proportion of missing values increases. We can also notice that the MAE values in Table IV are nearly double of the MAE when there are no missing values, listed in Table II, which means the missing values really affect the performance of the model. In conclusion, the model's capability of dealing with missing values is acceptable, but there are still rooms for us to improve the model robustness.

### C. Model Scalability

All above experiments are conducted based on freeway traffic state data, which does not cover urban roadways. To test the scalability of the proposed approach, we adopt the INRIX data, which covers a wide range of roadway segments, including freeways, arterials, urban streets, and even ramps, to predict the speed of the whole traffic network in the Seattle downtown area. The INRIX traffic network is shown in Fig. 6 (b). This prediction task is more challenging, not only because the traffic network consists of multiple types of roads, also due to the speed limits of these roads varies from 20 mph to 60 mph.

TABLE IV
PERFORMANCES COMPARISON OF THE SBU-LSTM WITH DIFFERENT
PROPORTIONS OF MISSING VALUES

| Proportion of missing values in input data | MAE | MAPE |
|---|---|---|
| 5% | 3.828 | 9.053 |
| 10% | 4.294 | 9.968 |
| 15% | 4.328 | 10.348 |
| 20% | 4.765 | 11.118 |
| 30% | 4.962 | 11.507 |

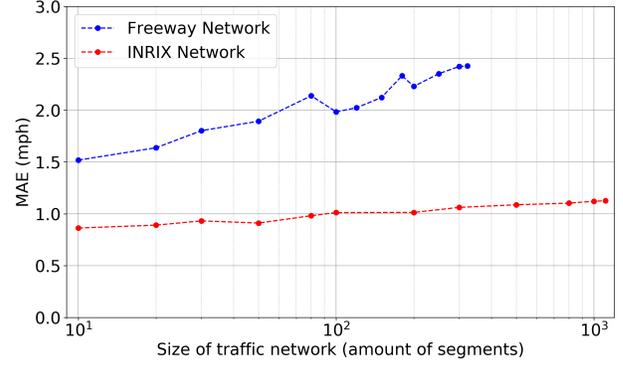

Fig. 9 Prediction MAE of the SBU-LSTM model versus traffic network size for both freeway and INRIX network. The x-axis is exponentially scaled to show the results for both datasets.

In this section, we use the basic SBU-LSTM model to predict speed for the INRIX traffic network containing more than 1000 roadway segments. The prediction MAE and MAPE is 1.126 mph and 4.212%, respectively, which is better than the that of the experiments based on loop detector data. This implies that the variation of the INRIX speed data might be relatively small. Further, the scalability of the model turns out to be quite remarkable, when we scale the size of traffic network. Fig. 9 shows the MAEs when applying the model to size-varying freeway and INRIX traffic networks. It shows that, when the size of the IRNIX traffic network increases, the prediction MAE increases slightly. In addition, size of freeway network also does not affect the prediction performance that much, considering the horizonal axis of Fig. 9 is an exponentially scaled. Hence, it is proved that the proposed approach is able to deal with multiple types of traffic network and works pretty good when the size of traffic network changes.

### D. Visualization and Potential Applications

Besides theoretical contribution, the proposed model has potential impact on the traffic speed prediction related applications. The proposed model and its visualized predicting results will soon be implemented on an extended version of a transportation data analytics platform [29, 30], which mainly utilizes artificial intelligence methods to solve transportation problems. The model's predicted results and the corresponding visualized traffic networks, like the studied freeway and INRIX traffic networks, shown in Fig. 10, will be public accessible via the platform.

It has been proved that network-wide prediction accuracy is high in previous sections. By investigating the predicted traffic speed for single locations, we find that the prediction performance is also very good and the trends of predicted and true values are pretty similar. For an example, Fig. 11 (a) and Fig. 11 (b) show the predicted and true speed values at two randomly selected locations from the freeway and INRIX traffic networks, respectively, during stochastically selected weekdays. It is obvious that recurring congestions at morning and evening peak hours are successfully predicted by the proposed approach. While, non-recurring congestions cannot be easily predicted by models without enough training data, like weather data, event data and incident data. The red box in Fig. 11 (c) tagged a sudden congestion in the late evening at the 114-04200 TMC place on the INRIX network, which is highly possible to be a non-recurring congestion. Thus, by feeding more data sources, like weather data and incident data, to the proposed model, distinguishing recurring and non-recurring congestions may be another applicable scenario soon.

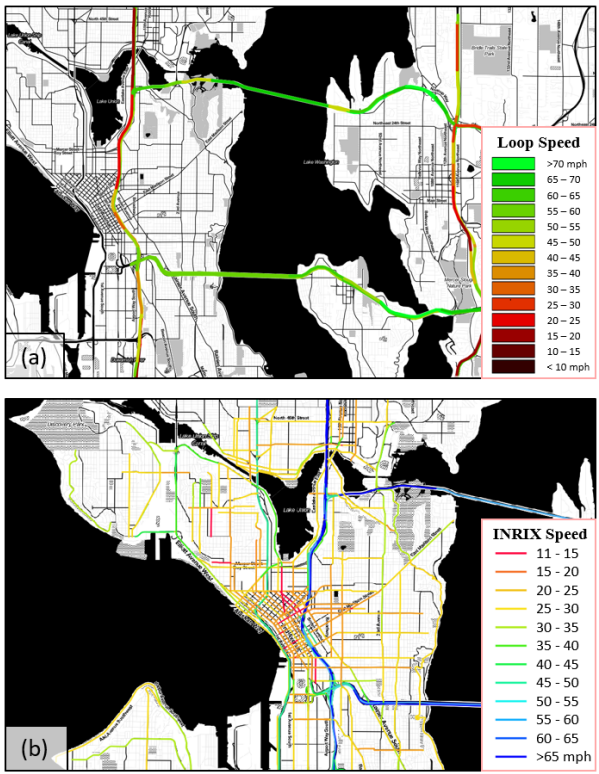

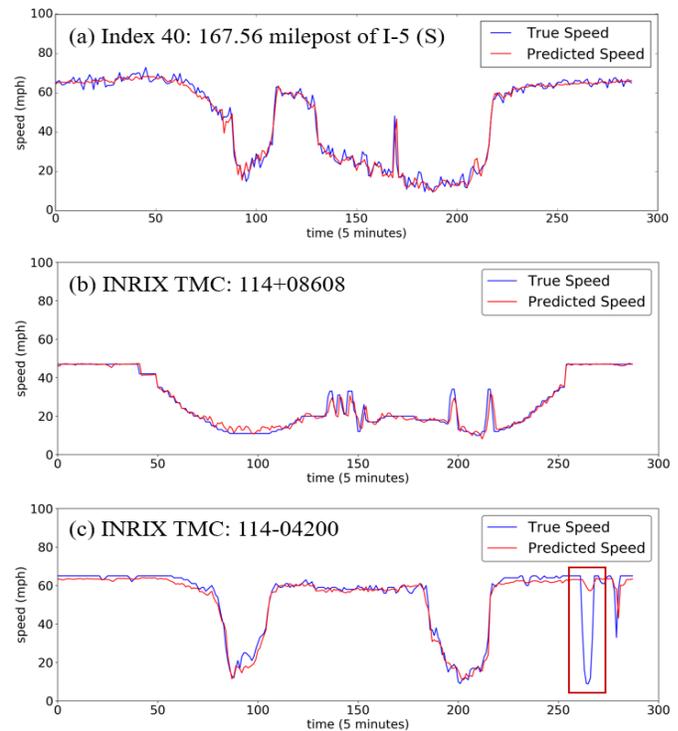

Fig. 10 Visualization of predicted traffic speed on freeway and INRIX traffic networks. (a) Freeway traffic network. (b) INRIX traffic network.

Fig. 11 Prediction Performance for single locations on a randomly selected weekday: (a) a sensor station at 167.56 milepost on I-5, (b) a TMC in the INRIX network, and (c) a TMC in the INRIX network, at which a non-recurring congestion occurred. X-axis represents the time of a day.

## IV. CONCLUSION AND FUTURE WORK

A deep stacked bidirectional and unidirectional LSTM neural network is proposed in this paper for network-wide traffic speed prediction. The improvements and contributions in this study mainly focus on four aspects: 1) we expand the traffic forecasting area to the whole traffic network, including freeway and urban traffic networks; 2) we propose a deep architecture stacked architecture considering both forward and backward dependencies of network-wide traffic data; 3) multiple influential factors for the proposed model are detailly analysed; and 4) a masking mechanism is adopted to handle missing values.

Experiment results indicate that the two-layers SBU-LSTM without middle layers is the best structure for network-wide traffic speed prediction. Comparing to LSTM, BDLSTM and other LSTM-based methods, the structure of stacking BDLSM and LSTM layers turns out to be more efficient to learn spatial-temporal features from the dataset. If the number of time lags of historical data is not large enough, prediction performance may decrease. The spatial order of input data and the spatial dimension of weight matrices in the last layer of the model almost has no influence on the prediction results. Additional information, like volume and occupancy, cannot significantly improve the predictive performance. Further, it is proved that the proposed model is suitable for predicting traffic speed on different types of traffic network.

Further improvements and extensions can be made based on this study. The model will be improved towards graph-based structure to learn and interpret spatial features. The model will be implemented on an artificial intelligence based transportation analytical platform. Potential applications, like non-recurring congestion detection, will be explored by combining other datasets.